\documentclass[runningheads]{llncs}

\usepackage[utf8]{inputenc}
\usepackage[T1,T2A]{fontenc}
\usepackage[russian,english]{babel}

\usepackage{graphicx}

\usepackage{url}

\usepackage{multirow}

\usepackage{adjustbox}

\usepackage{float}

\usepackage{makecell}

\begin{document}
\title{Demographic and Linguistic Bias Evaluation in Omnimodal Language Models}
\titlerunning{Demographic and Linguistic Bias Evaluation in Omnimodal LMs}

\author{Alaa Elobaid\inst{1}\orcidID{0009-0009-2399-9754}}

\authorrunning{A. Elobaid}

\institute{Freie Universit\"at Berlin, Berlin, Germany \\
\email{alaa.elobaid@fu-berlin.de}}

\maketitle              

\begin{abstract}
This paper provides a comprehensive evaluation of demographic and linguistic biases in omnimodal language models that process text, images, audio, and video within a single framework. Although these models are being widely deployed, their performance across different demographic groups and modalities is not well studied. Four omnimodal models are evaluated on tasks that include demographic attribute estimation, identity verification, activity recognition, multilingual speech transcription, and language identification. Accuracy differences are measured across age, gender, skin tone, language, and country of origin. The results show that image and video understanding tasks generally exhibit better performance with smaller demographic disparities. In contrast, audio understanding tasks exhibit significantly lower performance and substantial bias, including large accuracy differences across age groups, genders, and languages, and frequent prediction collapse toward narrow categories. These findings highlight the importance of evaluating fairness across all supported modalities as omnimodal language models are increasingly used in real-world applications.

\end{abstract}

\section{Introduction}
Omnimodal Language Models (OLM) are a subset of Multimodal Language Models (MLM) that can process more than two modalities simultaneously. In contrast to unimodal Language Models (LM) that process exclusively textual input, or bimodal LMs such as Vision Language Models (VLM) (text and vision) and audio LMs (text and audio), OLMs are capable of processing three or more modalities including text, vision, and audio within a unified framework. The term "omnimodal" was popularized by OpenAI with the release of GPT-4o \cite{hurst2024gpt} and has since been adopted by Amazon with Nova 2 Omni \cite{novaomni2025} and Alibaba with Qwen2.5-Omni \cite{xu2025qwen25omni}. 

However, bias evaluation research in MLMs has primarily focused on evaluating bimodal models such as VLMs and Audio LMs on a single modality \cite{narayan2025facexbench,perera2025investigating,kulkarni2025unveiling,nakatumba-nabende_systematic_2025}.

This paper aims to evaluate the demographic bias in the form of accuracy disparities between demographic groups such as age, skin colour, and gender in four OLMs by quantifying it using a combination of visual and audio tasks: demographic attribute estimation from both image and audio, multilingual speech transcription and language identification, image-based identity verification, and activity recognition from video.

\section{Related work}
MLMs have evolved from modality specific (bimodal) models to omnimodal models \cite{jiang-etal-2025-specific} capable of processing combinations of visual, auditory, textual, and other modalities within unified frameworks. While recent research has addressed technical challenges in OLMs such as modality bias, where models disproportionately attend to dominant modalities due to training data imbalances \cite{chen2024quantifying}, demographic bias has received far less attention. Existing demographic bias evaluation work in MLMs has focused predominantly on vision language models, examining social biases across gender, race, and age \cite{narayan2025facexbench,perera2025investigating,cheng2024social,huang2025visbias}. However, comparable evaluation for OLMs, particularly those with audio processing capabilities, is largely missing. This gap is critical as OLMs capable of processing both vision and audio are increasingly deployed in real-world applications such as multimodal age verification \cite{biometricupdate2024multimodal} and multimodal healthcare assistants  \cite{info:doi/10.2196/59505} without a clear understanding of how they perform across demographic groups.

While this study focuses on OLMs that process multiple modalities simultaneously, it is important to distinguish these from video LMs such as LLaVA-Video \cite{zhang2024video}. Video LMs employ explicit temporal modeling mechanisms including temporal attention modules and embeddings to capture inter-frame relationships, prioritizing depth of temporal visual understanding. However, when these video specialized models are extended to process audio, a recent study shows that they suffer from cross modal hallucinations and do not support the level of audio understanding needed for nuanced audio processing tasks \cite{sung2024avhbench}. In contrast, OLMs are architecturally designed with native audio encoders that enable nuanced audio understanding tasks including multilingual speech transcription, speaker identification, and acoustic scene analysis alongside visual processing. Therefore, they are the focus of this work.

The remainder of this literature review examines existing research on demographic bias in VLMs and traditional automated speech recognition systems, establishing the foundation for understanding bias in OLMs.

Narayan et al. \cite{narayan2025facexbench} created the FaceXBench benchmark and evaluated 26 open-source and 2 proprietary VLMs on 14 face understanding tasks including demographic attribute estimation such as age, race, and gender. However, formulating bias and fairness solely as accuracy in demographic attribute estimation ignores how the model's performance varies across different demographic groups. This  age estimation, gender prediction, and race estimation approach to bias assessment was also adopted by Shahreza et al. \cite{shahreza2025facellm}, who developed a specialized VLM for face understanding tasks and evaluated it on the same FaceXBench benchmark \cite{narayan2025facexbench}.

Perera et al. \cite{perera2025investigating} address this limitation by developing a benchmark of 10,000 VQA-based questions for attribute estimation and reporting model  performance across demographic groups. Nonetheless, their study maintains the same single-task formulation of bias. Beyond demographic prediction accuracy, some works have investigated social representation bias in how VLMs portray different demographic groups \cite{cheng2024social,huang2025visbias}, though such qualitative bias assessment falls outside the scope of this study, which focuses on quantifiable performance disparities across demographic groups.

Bias in Automated Speech Recognition (ASR) is a well-documented phenomenon across traditional ASR models such as Whisper \cite{radford2022whisper}, wav2vec 2.0 \cite{baevski2020wav2vec}, and Massively Multilingual Speech (MMS) \cite{pratap2023scaling}. Early work by Feng et al. \cite{feng_quantifying_2021} quantified bias in Dutch ASR systems across gender, age, regional accents, and non-native accents through phoneme-level error analysis. Subsequent research has included diverse linguistic contexts, with studies examining Portuguese \cite{kulkarni_balancing_2024}, Italian dialects \cite{shim_dialetto_2024}, and African low-resource languages \cite{nakatumba-nabende_systematic_2025,imam_automatic_2025}. Research has consistently revealed performance disparities, with minority dialects and non-native speakers experiencing higher Word Error Rates (WER) compared to standard language counterparts \cite{rio_accents_2023,serditova_automatic_2025,torgbi_adapting_2025}.

However, despite extensive bias quantification research in ASR-specific models, bias in MLMs with audio processing capabilities remains largely unexplored. Additionally, integrating audio processing capabilities into MLMs introduces bias manifestations that are fundamentally different from those in traditional ASR-specific models. For example, the phoneme-level error analysis methodology that has been central to traditional ASR bias research \cite{feng_towards_2024} is inapplicable to MLMs. Unlike traditional ASR architectures, MLMs operate as end-to-end systems that process audio through learned latent representations without explicit phoneme-level outputs, making fine-grained linguistic analysis unachievable. As a result, bias evaluation in OLMs and other audio-capable LMs must focus on downstream task performance rather than intermediate linguistic representations, relying on higher-level metrics such as WER or Character Error Rate for ASR tasks. Despite this limitation, audio-capable LMs offer unique zero-shot speech understanding capabilities beyond traditional speech recognition \cite{wang2024comprehensive}, enabling tasks such as demographic attribute classification, language identification, and topic recognition without task-specific training.

\section{Experimental setup}
Four representative OLMs are evaluated on four benchmark datasets spanning image, audio, and video modalities. The selected datasets provide comprehensive demographic attribute estimation, face verification, speech recognition, language identification, and activity recognition tasks, while enabling evaluation across diverse demographic groups and linguistic contexts.
\subsection{Datasets}

To test the models across all three modalities and the devised tasks, the following four datasets are selected:

\textbf{Casual Conversations V2 (CC2) \cite{porgali2023cc2}} is a large dataset for benchmarking multimodal AI systems on fairness and robustness. It contains 26,467 videos from 5,567 participants across seven countries, totaling 674 hours. The dataset provides extensive annotations, including age, gender, language, geo-location, skin tone, activity, and audio transcriptions. Videos consist of either scripted readings from Dostoevsky's \textit{The Idiot} or nonscripted responses to preset questions. This breadth enables diverse tasks such as demographic attribute estimation, activity recognition, language identification and ASR across multiple languages.

\textbf{Casual Conversations V1 (CC1) \cite{hazirbas2021cc1}} consists of 45,186 videos from 3,011 participants, with an average video length of approximately 1 minute totaling 846 hours of content. The dataset is annotated for age, gender, skin type, and lighting conditions. CC1 mainly involves English-language videos and is targeted at ASR and demographic attribute estimation tasks in vision and audio modalities.

\textbf{Balanced Faces in the Wild (BFW) \cite{robinson2020bfw}} is a dataset comprising 20,000 facial images of 800 identities, annotated for race, age, and gender. BFW focuses on facial ID verification and demographic attribute estimation in images.

\textbf{Mozilla Common Voice (MCV) \cite{ardila2019common}} is a large-scale multilingual speech dataset maintained by Mozilla. MCV 22.0 is used, which includes 33,815 hours of speech recordings from thousands of contributors across 137 languages, with annotations for transcription, language, age, and gender. Audio clips span 1 to 15 seconds. It is widely used for benchmarking ASR and language identification.

\subsection{Tasks}

\textbf{Image-Based Tasks:}
Models are evaluated on seven image understanding tasks using the CC1, CC2, and BFW datasets. Age classification requires models to predict age as an integer value, with accuracy computed using a ±5 year tolerance during evaluation, acknowledging the inherent difficulty of exact age prediction from visual appearance alone. Fitzpatrick skin tone classification also employs a ±1 category tolerance during evaluation due to the challenging nature of skin type classification, where even trained dermatologists report only moderate agreement \cite{groh2022transparency}. Gender classification and country prediction evaluate a model’s ability to infer binary gender and country of origin from faces. Face verification, evaluated on the BFW dataset, tests the model's ability to determine whether two facial images belong to the same person.

\textbf{Video-Based Tasks: }
Video tasks assess models' understanding of human actions and body visibility using the CC2 dataset. Action classification requires models to identify the person's physical activity from six categories: rotating, standing, sitting, walking, laying, or waving. Visibility classification evaluates the model's ability to determine body visibility from four categories: only head visible, upper body visible, full body visible, or lower body visible. 

To assess temporal understanding, evaluation clips that include behavior transitions are constructed. For each action or visibility annotated segment, an adjacent segment of equal length with a different label is identified and concatenated into a single clip. This approach tests whether models can correctly distinguish when behaviors change over time.

\textbf{Audio-Based Tasks: }
Audio tasks evaluate the models' multilingual speech understanding capabilities. Age and gender classification from audio requires models to predict those attributes solely from voice characteristics (with a ±5 year tolerance for age). Language identification tests the model's ability to recognize the language spoken in audio across multiple languages including English, Spanish, Portuguese, Hindi, Tagalog, Indonesian, Telugu, Tamil, and Vietnamese. Speech transcription evaluates ASR capabilities using Word Accuracy (WA) as the primary metric, which measures the proportion of words correctly recognized relative to a reference transcript. Word Accuracy is calculated as

\[
WA = 100 \times \left( 1 - \frac{S + D + I}{N} \right)
\]
where \(S\) is the number of substitutions, \(D\) the deletions, \(I\) the insertions, and \(N\) the total words in the reference transcript.

\subsection{Models}
\label{models:sec}

\textbf{Gemini 2.5 Flash (proprietary) \cite{comanici2025gemini}}
is an efficient model from Google's Gemini family of highly capable multimodal large language models, optimized for low-latency reasoning across text, images, audio, and video. It features a native multimodal encoder trained end-to-end for unified omnimodal understanding and rapid generation.

\textbf{Gemma 3n (open-weights) \cite{gemma2025team}}
is an efficient MLM developed by Google DeepMind that processes both visual and auditory inputs alongside text. The model employs a MobileNet-V5-300M encoder for visual feature extraction. For audio processing, Gemma 3n uses a 0.68B parameter encoder based on the Universal Speech Model (USM) conformer-based architecture. The E2B configuration is selected for evaluation, with 5.44B total and 1.91B effective parameters.
\textbf{Qwen 2.5 Omni (open-weights) \cite{xu2025qwen25omni}}
is a MLM that integrates vision, audio, and text modalities in an end-to-end architecture. For visual encoding, the model utilizes an adapted version of the Qwen2.5-VL vision encoder, while audio processing is handled by a modified Whisper-large-v3 encoder with 1.55B parameters. This study evaluates the 3B parameter variant, which offers a balance between performance and efficiency for multimodal understanding tasks.

\textbf{Phi-4 Multimodal (open-weights) \cite{abouelenin2025phi4}} is Microsoft's MLM that processes images and audio in addition to text using a mixture-of-LoRAs architecture. The model incorporates SigLIP-400M for image encoding, and employs a custom Conformer with 460M parameters for audio feature extraction. Built on the Phi-4-Mini base, the complete Phi-4 Multimodal model totals 5.6B parameters.

\subsection{Evaluation Protocol}

All models are evaluated using structured JSON prompts that specify the expected output format with prediction-confidence pairs for each task. Full-size original images and audio files are passed to the models without preprocessing. Complete prompts are provided in Appendix~\ref{sec:prompts}.

For categorical attributes (e.g., gender, language, country), prompts provide an enumerated list of valid discrete values from which models must select. For continuous attributes (e.g., age), prompts request integer predictions.

Prompts constrain model outputs to response categories aligned with dataset annotations. For speech transcription tasks, prompts specify that transcribed text must use native script characters rather than romanized transliterations, ensuring proper evaluation of multilingual capabilities. Models are instructed to respond only with valid JSON, facilitating parsing and evaluation at scale.

\subsection{Reproducibility and null prediction handling}
To promote reproducibility, greedy decoding is used by setting the sampling temperature to 0 and top-p to 1, ensuring that the model consistently selects the highest-probability token at each generation step \cite{song2025good}. 

Null predictions are defined as outputs where the model fails to generate valid responses or returns explicit refusal indicators. Null predictions are included in the total sample count, effectively counting as incorrect predictions. Notably, null predictions occur exclusively in audio tasks.

\section{Results and discussion}
This section presents the demographic performance evaluation of the four OLMs presented in Section \ref{models:sec} across image, video, and audio understanding tasks. Accuracy is reported as the primary performance metric and bias is quantified as the standard deviation of accuracy across demographic groups. For each demographic group, 120 samples are used for audio-based and image-based tasks, while 20 samples are used for video-based tasks.

\subsection{Overall Accuracy}
\label{subsec:overall-accuracy}

\begin{table}[H]
\centering
\caption{Accuracy Across All Tasks, Models, and Datasets (All values are percentages)}
\label{tab:multi_dataset_results}
\footnotesize
\adjustbox{width=\textwidth,center}{
\begin{tabular}{|l|l|l|l|c|c|c|c|}
\hline
\textbf{Task Category} & \textbf{Task} & \textbf{Dataset} & \textbf{Metric} & \textbf{Gemini} & \textbf{Phi} & \textbf{Qwen} & \textbf{Gemma} \\
\hline
\multirow{5}{*}{\textbf{Image-Based}} 
& \multirow{2}{*}{Age Classification} & CC1 & Accuracy & 54.2 & 32.5 & 38.5 & 51.2 \\
\cline{3-8}
& & CC2 & Accuracy & 63.1 & 42.8 & 46.5 & 52.5 \\
\cline{2-8}
& \multirow{2}{*}{Fitzpatrick skin} & CC1 & Accuracy & 90.4 & 53.9 & 51.2 & 44.3 \\
\cline{3-8}
& & CC2 & Accuracy & 87.8 & 49.9 & 50.0 & 43.4 \\
\cline{2-8}
& \multirow{2}{*}{Gender Classification} & CC1 & Accuracy & 96.7 & 95.4 & 95.9 & 98.8 \\
\cline{3-8}
& & CC2 & Accuracy & 97.5 & 48.8 & 95.0 & 94.4 \\
\cline{2-8}
& \multirow{1}{*}{Country Prediction} & CC2 & Accuracy & 83.2 & 17.5 & 38.9 & 35.9 \\
\cline{2-8}
& \multirow{1}{*}{Face Verification} & BFW & Accuracy & 90.0 & 69.5 & 70.5 & 75.8 \\
\hline
\textbf{Video-Based} & Action & CC2 & Accuracy & 84.2 & 61.6 & 76.4 & 77.9 \\
\cline{2-8}
& \multirow{1}{*}{Visibility} & CC2 & Accuracy & 51.8 & 50.3 & 48.2 & 53.2 \\
\hline
\multirow{7}{*}{\textbf{Audio-Based}} 
& \multirow{2}{*}{Age Classification} & CC2 & Accuracy & 36.0 & 25.6 & 30.6 & 36.4 \\
\cline{3-8}
& & MCV & Accuracy & 26.4 & 19.4 & 43.8 & 23.6 \\
\cline{2-8}
& \multirow{2}{*}{Gender Classification} & CC2 & Accuracy & 99.6 & 55.5 & 91.5 & 80.9 \\
\cline{3-8}
& & MCV & Accuracy & 87.5 & 21.9 & 88.7 & 37.9 \\
\cline{2-8}
& \multirow{2}{*}{Language ID} & CC2 & Accuracy & 99.7 & 49.4 & 87.3 & 95.1 \\
\cline{3-8}
& & MCV & Accuracy & 97.1 & 100 & 100 & 100 \\
\cline{2-8}
& \multirow{2}{*}{Speech Transcription} & CC2 & WA & 83.97 & 34.86 & 43.57 & 69.35 \\
\cline{3-8}
& & MCV & WA & 72.15 & -5.81 & 52.12 & 40.40 \\
\hline
\end{tabular}
}
\end{table}

Models demonstrate variable performance across image tasks, with scores generally 
exceeding 50\% and reaching as high as 98.8\% for gender classification. Although task-specific baseline results are not included, the relative differences 
in model performance across tasks, datasets, and modalities provide insights into task difficulty and model behaviour. Gemini leads in most image tasks, achieving the highest scores in Fitzpatrick skin classification (87.8\% on CC2), gender classification (97.5\% on CC2), country prediction (83.2\%), and face verification (90.0\%). Gemma achieves results comparable to Gemini across age and gender 
classification, and face verification. The CC2 dataset consistently yields better results than CC1 for comparable tasks, suggesting dataset quality differences. Gender classification is the easiest vision task with all models except Phi consistently achieving over 92\% accuracy, while country prediction on CC2 proves challenging with scores ranging from 17.5\% to 83.2\%. For video tasks, Gemini maintains the highest performance in action recognition (84.2\%), while visibility classification proves challenging for all models with accuracies around 50\%.

In contrast to vision tasks, audio tasks reveal significantly weaker results across models, with most scores falling below 50\%. Gemini demonstrates substantially stronger audio understanding capabilities compared to other models, achieving near-perfect language identification (99.7\% on CC2, 97.1\% on MCV), high gender classification accuracy (99.6\% on CC2, 87.5\% on MCV), and the best speech transcription performance (83.97\% WA on CC2, 72.15\% on MCV). The dataset characteristics appear to strongly influence results, with Qwen, Gemma, and Gemini performing better on CC2's longer audio clips than MCV' short clips. However, Phi shows an opposite pattern, performing better on MCV for language identification (100\%), which likely indicates data leakage from its training on "20k hours selected public transcribed" data \cite{abouelenin2025phi4} that may have included MCV samples. Gender classification in audio reveals notable model performance differences. Qwen achieves consistent results across both datasets (91.5\% CC2, 88.7\% 
MCV), whereas Gemma achieves lower results on CC2 (80.9\%) with a notable decline 
on MCV (37.9\%). Phi's results consistently remain below 56\%.

\subsection{Image-Based Tasks}

\subsubsection{Face Verification - BFW (Table \ref{tab:face_verification_accuracy})} \

\begin{table}[h]
\centering
\caption{Face Verification Accuracy by Demographic Group}
\label{tab:face_verification_accuracy}
\footnotesize
\adjustbox{width=\textwidth,center}{
\begin{tabular}{|l|c|c|c|c|c|c|c|c|c|c|c|c|}
\hline
\textbf{Model} & \textbf{AF} & \textbf{AM} & \textbf{BF} & \textbf{BM} & \textbf{IF} & \textbf{IM} & \textbf{WF} & \textbf{WM} & \textbf{Avg} & \textbf{Std} & \textbf{Std (Race)} & \textbf{Std (Gender)} \\
\hline
Gemini & 80.1 & 91.1 & 94.5 & 87.5 & 87.9 & 93.0 & 91.5 & 94.5 & 90.0 & 4.5 & 2.7 & 1.5 \\
\hline
Gemma & 69.3 & 80.0 & 82.0 & 76.3 & 64.2 & 77.4 & 75.8 & 81.2 & 75.8 & 5.8 & 3.3 & 3.0 \\
\hline
Qwen & 58.0 & 64.0 & 76.0 & 73.0 & 74.5 & 77.5 & 73.0 & 68.0 & 70.5 & 6.2 & 5.8 & 0.1 \\
\hline
Phi4 & 66.0 & 75.0 & 72.5 & 61.5 & 55.0 & 71.5 & 73.5 & 81.0 & 69.5 & 7.7 & 5.2 & 2.8 \\
\hline
\end{tabular}
}
\\[0.5ex]
\scriptsize
All values are percentages. A: Asian. B: Black. I: Indian. W: White. F: Female. M: Male.
\end{table}

In face verification, Gemini demonstrates the highest performance (90.0\% average) with the lowest total Std (4.5\%). The model shows particularly strong performance for Black females (94.5\%) and White males (94.5\%), while Asian females represent the weakest group (80.1\%). The gender-based standard deviation is relatively low across models (ranging from 0.1\% to 3.0\%), with Qwen achieving near-parity (Std Gender: 0.1\%). Race-based disparities are more pronounced than gender disparities across all models, as evidenced by higher Std (Race) values compared to Std (Gender). Asian females consistently represent the lowest-performing demographic group across three models (80.1\% for Gemini, 69.3\% for Gemma, 58.0\% for Qwen). Phi exhibits the highest overall disparity (Std: 7.7\%) with particularly poor performance on Indian females (55.0\%) and Black males (61.5\%). Qwen shows the most severe race-based bias (Std Race: 5.8\%) but achieves near gender parity. Gemma demonstrates moderate bias with near-balanced race (3.3\%) and gender (3.0\%) standard deviations.

\begin{table}[h]
\centering
\caption{Per-Attribute Accuracy with Top Predictions (CC1, 120 samples per group)}
\label{tab:cc1-image}
\footnotesize
\adjustbox{width=\textwidth,center}{
\begin{tabular}{|l|l|c|c|c|c|c|c|c|c|c|c|c|c|}
\hline
\textbf{Attr.} & \textbf{Group} & \multicolumn{3}{c|}{\textbf{Phi}} & \multicolumn{3}{c|}{\textbf{Qwen}} & \multicolumn{3}{c|}{\textbf{Gemma}} & \multicolumn{3}{c|}{\textbf{Gemini}} \\
& & \textbf{Acc} & \textbf{Prd} & \textbf{Top} & \textbf{Acc} & \textbf{Prd} & \textbf{Top} & \textbf{Acc} & \textbf{Prd} & \textbf{Top} & \textbf{Acc} & \textbf{Prd} & \textbf{Top} \\
\hline
\multirow{8}{*}{Age} 
& 18--29 & 57.5 & 57.5 & 30(53) & 60.8 & 9.9 & 30(49) & 74.2 & 15.3 & 25(54) & 55.0 & 20.4 & 26(47.5) \\
& 30--39 & 43.3 & 43.3 & 30(68) & 66.7 & 36.2 & 30(68) & 64.2 & 26.0 & 35(39) & 48.3 & 11.8 & 43(23.3) \\
& 40--49 & 21.7 & 21.7 & 30(48) & 42.5 & 21.8 & 45(38) & 38.3 & 12.6 & 35(22) & 60.0 & 22.5 & 43(28.3) \\
& 50--59 & 37.5 & 37.5 & 50(45) & 28.3 & 10.0 & 45(50) & 53.3 & 19.3 & 55(39) & 55.8 & 17.2 & 54(20.0) \\
& 60--69 & 22.5 & 22.5 & 50(43) & 24.2 & 19.9 & 60(38) & 45.8 & 20.3 & 60(29) & 65.8 & 23.1 & 67(17.5) \\
& 70+ & 12.5 & 12.5 & 60(63) & 8.3 & 2.1 & 60(65) & 31.7 & 6.5 & 65(39) & 40.2 & 4.3 & 67(26.5) \\
\cline{2-14}
& \textit{Overall} & \textit{32.5} & \textit{--} & \textit{--} & \textit{38.5} & \textit{--} & \textit{--} & \textit{51.2} & \textit{--} & \textit{--} & \textit{54.2} & \textit{--} & \textit{--} \\
\cline{2-14}
& \textit{Std} & \textit{15.2} & \textit{--} & \textit{--} & \textit{22.4} & \textit{--} & \textit{--} & \textit{16.8} & \textit{--} & \textit{--} & \textit{8.2} & \textit{--} & \textit{--} \\
\hline
\multirow{4}{*}{Gender} 
& Female & 93.3 & 47.9 & F(93) & 94.2 & 48.0 & F(94) & 99.2 & 50.4 & F(99) & 94.2 & 47.7 & F(94.2) \\
& Male & 97.5 & 52.1 & M(98) & 97.5 & 52.0 & M(98) & 98.3 & 49.6 & M(98) & 99.2 & 52.3 & M(99.2) \\
\cline{2-14}
& \textit{Overall} & \textit{95.4} & \textit{--} & \textit{--} & \textit{95.9} & \textit{--} & \textit{--} & \textit{98.8} & \textit{--} & \textit{--} & \textit{96.7} & \textit{--} & \textit{--} \\
\cline{2-14}
& \textit{Std} & \textit{3.0} & \textit{--} & \textit{--} & \textit{2.3} & \textit{--} & \textit{--} & \textit{0.6} & \textit{--} & \textit{--} & \textit{2.5} & \textit{--} & \textit{--} \\
\hline
\multirow{8}{*}{\makecell[l]{Skin\\tone}} 
& 1 & 54.2 & 0.0 & 2(54) & 0.0 & 0.0 & 3(100) & 60.0 & 46.4 & 1(47) & 78.3 & 0.0 & 2(78.3) \\
& 2 & 100.0 & 29.6 & 3(60) & 100.0 & 0.0 & 3(100) & 98.3 & 11.5 & 1(59) & 90.8 & 28.9 & 2(63.3) \\
& 3 & 100.0 & 70.4 & 3(64) & 100.0 & 96.2 & 3(100) & 39.2 & 24.4 & 1(61) & 96.7 & 18.2 & 3(40.8) \\
& 4 & 66.7 & 0.0 & 3(67) & 100.0 & 3.1 & 3(98) & 31.7 & 17.6 & 1(57) & 96.7 & 27.8 & 4(74.2) \\
& 5 & 0.0 & 0.0 & 3(91) & 5.0 & 0.7 & 3(95) & 36.7 & 0.0 & 4(37) & 95.8 & 18.5 & 5(49.2) \\
& 6 & 0.0 & 0.0 & 3(95) & 2.5 & 0.0 & 3(85) & 0.0 & 0.0 & 4(58) & 84.2 & 6.7 & 5(52.5) \\
\cline{2-14}
& \textit{Overall} & \textit{53.9} & \textit{--} & \textit{--} & \textit{51.2} & \textit{--} & \textit{--} & \textit{44.3} & \textit{--} & \textit{--} & \textit{90.4} & \textit{--} & \textit{--} \\
\cline{2-14}
& \textit{Std} & \textit{42.8} & \textit{--} & \textit{--} & \textit{49.1} & \textit{--} & \textit{--} & \textit{35.2} & \textit{--} & \textit{--} & \textit{7.0} & \textit{--} & \textit{--} \\
\hline
\end{tabular}
}
\\[0.5ex]
\scriptsize
Acc: Accuracy (\%) with tolerance (±5 years for age, ±1 for skin tone). \\
Prd: Percentage of samples predicted as the attribute group. \\
Top: Most frequent prediction with percentage of samples receiving that prediction in brackets.
\end{table}

\subsubsection{CC1 Image (Table \ref{tab:cc1-image})} \

\textbf{Age classification}: Gemini has the lowest age bias (Std: 8.2\%) compared to Phi (15.2\%), Qwen (22.4\%), and Gemma (16.8\%). Open-source models exhibit systematic bias against older adults, with 70+ accuracy of only 8.3-31.7\%, while Gemini maintains 40.2-65.8\% across age groups. Qwen and Phi show extreme prediction concentration (36.2\% at 30-39 and 57\% at 18-29), whereas Gemini and Gemma distribute predictions more evenly (4.3-23.1\% and 6.5-26.0\%).

\textbf{Gender classification}: All models demonstrate minimal gender accuracy disparity (Std: 0.6-3.0\%), indicating balanced performance. However, for every model except Gemma, accuracy is higher for males by 2 to 5\%.

\textbf{Skin tone classification}: Qwen, Phi, and Gemma achieve 0-5\% accuracy for darker skin tones (Types 5-6). Qwen and Phi4 predict Type 3 for 96.2\% and 70.4\% of samples, respectively, while Gemma defaults to Type 1 with a 46.4\% prediction rate. Gemini's bias in the form of Std is much lower (Std: 7.0\%), maintaining 78.3-96.7\% accuracy across all Fitzpatrick types with more balanced predictions.
\begin{table}[h]
\centering
\caption{Per-Attribute Accuracy with Top Predictions (CC2, 120 samples per group)}
\label{tab:cc2-image}
\footnotesize
\adjustbox{width=\textwidth,center}{
\begin{tabular}{|l|l|c|c|c|c|c|c|c|c|c|c|c|c|}
\hline
\textbf{Attr.} & \textbf{Group} & \multicolumn{3}{c|}{\textbf{Phi}} & \multicolumn{3}{c|}{\textbf{Qwen}} & \multicolumn{3}{c|}{\textbf{Gemma}} & \multicolumn{3}{c|}{\textbf{Gemini}} \\
& & \textbf{Acc} & \textbf{Prd} & \textbf{Top} & \textbf{Acc} & \textbf{Prd} & \textbf{Top} & \textbf{Acc} & \textbf{Prd} & \textbf{Top} & \textbf{Acc} & \textbf{Prd} & \textbf{Top} \\
\hline
\multirow{6}{*}{Age} 
& 18--29 & 65.0 & 46.3 & 30(45) & 81.2 & 55.0 & 25(49) & 97.5 & 44.1 & 20(98) & 88.3 & 35.4 & 25(37) \\
& 30--39 & 52.5 & 75.0 & 30(75) & 66.2 & 78.7 & 30(64) & 45.0 & 28.4 & 20(51) & 40.0 & 24.2 & 28(24) \\
& 40--49 & 17.5 & 26.2 & 30(61) & 22.5 & 28.7 & 35(38) & 18.8 & 13.8 & 30(58) & 50.0 & 20.8 & 45(27) \\
& 50+ & 36.2 & 56.3 & 50(48) & 16.2 & 15.0 & 45(54) & 48.8 & 12.8 & 40(34) & 74.2 & 18.8 & 55(29) \\
\cline{2-14}
& \textit{Overall} & \textit{42.8} & \textit{--} & \textit{--} & \textit{46.5} & \textit{--} & \textit{--} & \textit{52.5} & \textit{--} & \textit{--} & \textit{63.1} & \textit{--} & \textit{--} \\
\cline{2-14}
& \textit{Std} & \textit{17.8} & \textit{--} & \textit{--} & \textit{27.8} & \textit{--} & \textit{--} & \textit{28.4} & \textit{--} & \textit{--} & \textit{19.1} & \textit{--} & \textit{--} \\
\hline
\multirow{4}{*}{Gender} 
& Female & 55.0 & 53.8 & F(51) & 93.8 & 48.8 & F(94) & 95.0 & 50.6 & F(95) & 98.3 & 50.8 & F(98) \\
& Male & 42.5 & 41.9 & F(56) & 96.2 & 51.2 & M(96) & 93.8 & 49.4 & M(94) & 96.7 & 48.8 & M(97) \\
\cline{2-14}
& \textit{Overall} & \textit{48.8} & \textit{--} & \textit{--} & \textit{95.0} & \textit{--} & \textit{--} & \textit{94.4} & \textit{--} & \textit{--} & \textit{97.5} & \textit{--} & \textit{--} \\
\cline{2-14}
& \textit{Std} & \textit{6.2} & \textit{--} & \textit{--} & \textit{1.2} & \textit{--} & \textit{--} & \textit{0.6} & \textit{--} & \textit{--} & \textit{0.8} & \textit{--} & \textit{--} \\
\hline
\multirow{8}{*}{Country} 
& Brazil & 13.8 & 13.8 & Ind(64) & 20.0 & 5.8 & USA(70) & 2.5 & 0.4 & USA(83) & 73.3 & 13.3 & Bra(73) \\
& India & 45.0 & 57.7 & Ind(45) & 26.2 & 5.0 & USA(68) & 88.8 & 37.7 & Ind(89) & 95.8 & 18.3 & Ind(96) \\
& Indonesia & 0.0 & 0.0 & Ind(61) & 51.2 & 14.4 & Ind(50) & 22.5 & 4.0 & Ind(65) & 78.3 & 13.5 & Ind(78) \\
& Mexico & 0.0 & 0.0 & Ind(68) & 23.8 & 4.6 & USA(75) & 5.0 & 1.2 & USA(88) & 71.7 & 12.6 & Mex(72) \\
& Philippines & 0.0 & 0.0 & Ind(56) & 13.8 & 2.3 & USA(46) & 7.5 & 0.8 & USA(50) & 89.2 & 18.9 & Phi(89) \\
& USA & 46.2 & 28.1 & Ind(53) & 98.8 & 67.3 & USA(99) & 88.8 & 55.4 & USA(89) & 90.8 & 21.2 & USA(91) \\
\cline{2-14}
& \textit{Overall} & \textit{17.5} & \textit{--} & \textit{--} & \textit{38.9} & \textit{--} & \textit{--} & \textit{35.9} & \textit{--} & \textit{--} & \textit{83.2} & \textit{--} & \textit{--} \\
\cline{2-14}
& \textit{Std} & \textit{20.5} & \textit{--} & \textit{--} & \textit{29.2} & \textit{--} & \textit{--} & \textit{38.0} & \textit{--} & \textit{--} & \textit{9.2} & \textit{--} & \textit{--} \\
\hline
\multirow{8}{*}{\makecell[l]{Skin\\tone}} 
& 1 & 3.8 & 0.0 & 4(78) & 0.0 & 0.0 & 3(99) & 8.8 & 23.1 & 3(90) & 62.5 & 0.0 & 2(63) \\
& 2 & 6.2 & 3.8 & 4(79) & 100 & 0.0 & 3(100) & 98.8 & 3.8 & 3(74) & 74.2 & 13.9 & 3(58) \\
& 3 & 82.5 & 1.5 & 4(75) & 100 & 99.8 & 3(100) & 78.8 & 60.8 & 3(71) & 100 & 24.7 & 4(55) \\
& 4 & 96.2 & 76.2 & 4(73) & 100 & 0.2 & 3(98) & 67.5 & 12.1 & 3(66) & 99.2 & 43.5 & 4(84) \\
& 5 & 96.2 & 18.5 & 4(73) & 0.0 & 0.0 & 3(100) & 6.2 & 0.2 & 1(53) & 97.5 & 9.9 & 4(86) \\
& 6 & 15.0 & 0.0 & 4(81) & 0.0 & 0.0 & 3(98) & 0.0 & 0.0 & 4(59) & 93.3 & 8.1 & 6(47) \\
\cline{2-14}
& \textit{Overall} & \textit{49.9} & \textit{--} & \textit{--} & \textit{50.0} & \textit{--} & \textit{--} & \textit{43.4} & \textit{--} & \textit{--} & \textit{87.8} & \textit{--} & \textit{--} \\
\cline{2-14}
& \textit{Std} & \textit{42.0} & \textit{--} & \textit{--} & \textit{50.0} & \textit{--} & \textit{--} & \textit{39.5} & \textit{--} & \textit{--} & \textit{14.3} & \textit{--} & \textit{--} \\
\hline
\end{tabular}
}
\\[0.5ex]
\scriptsize
Abbreviations as in Table \ref{tab:cc1-image}.
\end{table}

\subsubsection{CC2 Image (Table \ref{tab:cc2-image})} \

\textbf{Age classification}: Both Phi and Qwen predictions are concentrated in the 30-39 age range. Phi consistently defaults to age 30 for groups 18-49 (45-75\% of predictions) as observed in the Top column, while Qwen predictions are more diverse, they are similarly concentrated most at ages 30 and 45. Gemma predicts age 20 for 98\% of 18-29 samples and 51\% of 30-39 samples.  Gemini distributes predictions more evenly (18.8-35.4\% per group) and exhibits the same inverted age pattern observed in CC1: performing weakest on middle-age groups (40.0\% for ages 30-39) while maintaining stronger performance at extremes.

\textbf{Gender classification}: Gender bias remains minimal across all models, between 0.6\% and 6.2\%, with balanced prediction distributions. Phi4 demonstrates the highest inconsistency, misclassifying 56\% of male samples as female as observed in the Top column.

\textbf{Country prediction}: Phi and Qwen exhibit extreme prediction concentration, defaulting to "India" or "USA" for the majority of samples regardless of actual country of origin. Gemini maintains more consistent performance across countries (Std: 9.2\%) with balanced predictions (12.6-21.2\% per country), while other models show near-complete failure for underrepresented countries (Brazil, Mexico, Philippines: 0-23.8\%).

\textbf{Skin tone classification}: Qwen exhibits near-total prediction collapse, concentrating 99.8\% of all predictions on Type 3 , resulting in 0\% accuracy for Types 1, 5, and 6. Phi shows similar concentration toward Type 4 (76.2\% of predictions), achieving near-zero accuracy for Types 1 and 2 and partial failure for Type 6. Gemma exhibits strong Type 3 bias (60.8\% prediction rate) with failure at extremes (0\% accuracy for Type 6 and only 6.2\% to 8.8\% for Types 1 and 5). Gemini maintains more consistent performance (Std: 14.3\%), although higher than its 7.0\% Std on CC1, indicating dataset-specific variation.

\subsection{Video-Based Tasks - CC2 (Table \ref{video-cc2})}

\begin{table}[h]
\centering
\caption{Demographic Group Performance on Video Understanding Tasks (CC2)}
\label{video-cc2}
\footnotesize
\adjustbox{max width=\textwidth,center}{  
\begin{tabular}{|l|l|
                c|c|c|c|
                c|c|c|c|
                c|c|c|c|
                c|c|c|c|}
\hline
\textbf{Attr.} & \textbf{Group}
  & \multicolumn{4}{c|}{\textbf{Phi}}
  & \multicolumn{4}{c|}{\textbf{Qwen}}
  & \multicolumn{4}{c|}{\textbf{Gemma}}
  & \multicolumn{4}{c|}{\textbf{Gemini}} \\
\cline{3-18}
& &
  \multicolumn{2}{c|}{\textbf{Act}} & \multicolumn{2}{c|}{\textbf{Vis}} &
  \multicolumn{2}{c|}{\textbf{Act}} & \multicolumn{2}{c|}{\textbf{Vis}} &
  \multicolumn{2}{c|}{\textbf{Act}} & \multicolumn{2}{c|}{\textbf{Vis}} &
  \multicolumn{2}{c|}{\textbf{Act}} & \multicolumn{2}{c|}{\textbf{Vis}} \\
\cline{3-18}
& &
  \textbf{Acc} & \textbf{Top} & \textbf{Acc} & \textbf{Top} &
  \textbf{Acc} & \textbf{Top} & \textbf{Acc} & \textbf{Top} &
  \textbf{Acc} & \textbf{Top} & \textbf{Acc} & \textbf{Top} &
  \textbf{Acc} & \textbf{Top} & \textbf{Acc} & \textbf{Top} \\
\hline
\multirow{6}{*}{Age}
& 18--29
  & 61.8 & st(86) & 50.5 & fb(100)
  & 78.2 & st(81) & 50.7 & ub(100)
  & 80.5 & st(77) & 51.9 & fb(50)
  & 85.5 & st(68) & 51.4 & fb(100) \\
& 30--39
  & 59.2 & st(88) & 53.1 & fb(99)
  & 75.2 & st(82) & 46.3 & ub(100)
  & 76.5 & st(80) & 64.7 & ub(64)
  & 84.2 & st(72) & 53.8 & fb(98) \\
& 40--49
  & 63.1 & st(85) & 53.6 & fb(100)
  & 75.5 & st(83) & 46.4 & ub(100)
  & 77.4 & st(78) & 46.4 & ub(100)
  & 88.2 & st(66) & 50.0 & ub(96) \\
& 50+
  & 61.7 & st(88) & -- & --
  & 76.5 & st(82) & -- & --
  & 78.3 & st(76) & -- & --
  & 83.0 & st(70) & -- & -- \\
\cline{2-18}
& \textit{Overall}
  & \textit{61.5} & -- & \textit{52.4} & -- 
  & \textit{76.3} & -- & \textit{47.8} & --
  & \textit{78.2} & -- & \textit{54.3} & --
  & \textit{85.2} & -- & \textit{51.7} & -- \\
\cline{2-18}
& \textit{Std}
  & \textit{1.4} & -- & \textit{1.4} & -- 
  & \textit{1.2} & -- & \textit{1.9} & --
  & \textit{1.3} & -- & \textit{7.5} & --
  & \textit{2.1} & -- & \textit{1.6} & -- \\
\hline
\multirow{4}{*}{Gender}
& Female
  & 63.4 & st(87) & 51.4 & fb(100)
  & 76.2 & st(82) & 48.7 & ub(100)
  & 78.2 & st(77) & 49.5 & ub(90)
  & 85.3 & st(70) & 52.6 & fb(79) \\
& Male
  & 58.9 & st(88) & 47.1 & fb(97)
  & 77.4 & st(81) & 48.1 & ub(100)
  & 76.5 & st(79) & 54.8 & fb(57)
  & 82.6 & st(70) & 50.4 & fb(100) \\
\cline{2-18}
& \textit{Overall}
  & \textit{61.1} & -- & \textit{49.2} & --
  & \textit{76.8} & -- & \textit{48.4} & --
  & \textit{77.3} & -- & \textit{52.1} & --
  & \textit{83.9} & -- & \textit{51.5} & -- \\
\cline{2-18}
& \textit{Std}
  & \textit{2.3} & -- & \textit{2.2} & --
  & \textit{0.6} & -- & \textit{0.3} & --
  & \textit{0.9} & -- & \textit{2.7} & --
  & \textit{1.4} & -- & \textit{1.1} & -- \\
\hline
\multirow{5}{*}{\makecell[l]{Skin\\tone}}
& I--II
  & 64.3 & st(83) & 49.7 & fb(100)
  & 74.6 & st(83) & 51.2 & ub(100)
  & 78.7 & st(77) & 52.1 & ub(98)
  & 80.7 & st(71) & 47.8 & fb(75) \\
& III--IV
  & 61.3 & st(88) & 48.4 & fb(92)
  & 79.3 & st(81) & 45.0 & ub(100)
  & 77.0 & st(78) & 54.1 & ub(64)
  & 83.7 & st(71) & 59.1 & fb(98) \\
& V--VI
  & 60.9 & st(86) & 48.9 & fb(100)
  & 75.4 & st(82) & 49.5 & ub(100)
  & 78.3 & st(77) & 52.0 & ub(59)
  & 85.1 & st(68) & 49.6 & fb(100) \\
\cline{2-18}
& \textit{Overall}
  & \textit{62.2} & -- & \textit{49.0} & --
  & \textit{76.4} & -- & \textit{48.6} & --
  & \textit{78.0} & -- & \textit{52.7} & --
  & \textit{83.2} & -- & \textit{52.2} & -- \\
\cline{2-18}
& \textit{Std}
  & \textit{1.5} & -- & \textit{0.5} & --
  & \textit{2.0} & -- & \textit{2.6} & --
  & \textit{0.7} & -- & \textit{0.9} & --
  & \textit{1.8} & -- & \textit{5.1} & -- \\
\hline
\end{tabular}}
\\[0.5ex]
\scriptsize
Acc: Accuracy (\%). Act: Action. Vis: Visibility. st: standing. fb: full body. ub: upper body. \\  
Top: Most frequent prediction with percentage of samples receiving that prediction in brackets.
\end{table}

\textbf{Action recognition: }In action recognition, models maintain low accuracy disparities with all standard deviations being below 2.3\% across all demographic groups. However, models exhibit strong prediction concentration toward "standing" (65.6\%–88.3\% of predictions) reflected in their moderate accuracies.

\textbf{Visibility classification:} Visibility classification shows comparably low disparities across all groups (Std: 0.3\%–7.5\%), though Gemma exhibits relatively high age-based variance (7.5\%) compared to other models (1.4\%–1.9\%). Overall accuracy per group remains low across models (47.8\%–54.3\%), partially attributed to overlapping category definitions where "upper body visible" and "full body visible" are not mutually exclusive. The absence of visibility transitions in 50+ age group samples prevents complete bias assessment across age groups.

\subsection{Audio-Based tasks}

\subsubsection{MCV Audio (Table \ref{tab:MCV-audio})}  \

\begin{table}[h]
\centering
\caption{Per-Attribute Accuracy with Top Predictions and Null Values (MCV, 120 samples per group, ±5 year tolerance for age)}
\label{tab:MCV-audio}
\footnotesize
\adjustbox{width=\textwidth,center}{
\begin{tabular}{|l|l|c|c|c|c|c|c|c|c|c|c|c|c|c|c|c|c|}
\hline
\textbf{Attr.} & \textbf{Group} & \multicolumn{4}{c|}{\textbf{Phi}} & \multicolumn{4}{c|}{\textbf{Qwen}} & \multicolumn{4}{c|}{\textbf{Gemma}} & \multicolumn{4}{c|}{\textbf{Gemini}} \\
& & \textbf{Acc} & \textbf{Prd} & \textbf{Nul} & \textbf{Top} & \textbf{Acc} & \textbf{Prd} & \textbf{Nul} & \textbf{Top} & \textbf{Acc} & \textbf{Prd} & \textbf{Nul} & \textbf{Top} & \textbf{Acc} & \textbf{Prd} & \textbf{Nul} & \textbf{Top} \\
\hline
\multirow{6}{*}{Age} 
& 18--29 & 43.3 & 1.2 & 41.7 & 30(43) & 77.5 & 15.8 & 0.0 & 30(40) & 67.5 & 37.1 & 16.7 & 25(34) & 25.0 & 2.1 & 0.0 & 35(47) \\
& 30--39 & 53.3 & 46.0 & 39.2 & 30(46) & 69.2 & 53.8 & 0.0 & 30(48) & 45.8 & 41.9 & 17.5 & 25(35) & 90.0 & 69.0 & 0.0 & 35(50) \\
& 40--49 & 6.7 & 9.6 & 44.2 & 30(47) & 28.3 & 20.6 & 0.0 & 30(43) & 5.0 & 3.5 & 11.7 & 25(36) & 31.7 & 28.3 & 0.0 & 35(47) \\
& 50+ & 1.7 & 1.7 & 38.3 & 30(43) & 3.3 & 1.2 & 0.0 & 30(37) & 1.7 & 1.7 & 6.7 & 25(33) & 0.0 & 0.4 & 0.0 & 35(45) \\
\cline{2-18}
& \textit{Overall} & \textit{19.4} & \textit{--} & \textit{--} & \textit{--} & \textit{43.8} & \textit{--} & \textit{--} & \textit{--} & \textit{23.6} & \textit{--} & \textit{--} & \textit{--} & \textit{26.4} & \textit{--} & \textit{--} & \textit{--} \\
\cline{2-18}
& \textit{Std} & \textit{22.4} & \textit{--} & \textit{--} & \textit{--} & \textit{30.2} & \textit{--} & \textit{--} & \textit{--} & \textit{28.5} & \textit{--} & \textit{--} & \textit{--} & \textit{35.8} & \textit{--} & \textit{--} & \textit{--} \\
\hline
\multirow{4}{*}{Gender} 
& Female & 58.3 & 62.5 & 40.0 & F(58) & 90.8 & 53.3 & 0.0 & F(91) & 75.0 & 73.3 & 10.8 & F(75) & 80.0 & 45.0 & 0.0 & F(80) \\
& Male & 0.0 & 0.8 & 32.5 & F(67) & 84.2 & 46.7 & 0.0 & M(84) & 17.5 & 15.8 & 10.8 & F(72) & 90.0 & 55.0 & 0.0 & M(90) \\
\cline{2-18}
& \textit{Overall} & \textit{21.9} & \textit{--} & \textit{--} & \textit{--} & \textit{88.7} & \textit{--} & \textit{--} & \textit{--} & \textit{37.9} & \textit{--} & \textit{--} & \textit{--} & \textit{87.5} & \textit{--} & \textit{--} & \textit{--} \\
\cline{2-18}
& \textit{Std} & \textit{29.1} & \textit{--} & \textit{--} & \textit{--} & \textit{3.3} & \textit{--} & \textit{--} & \textit{--} & \textit{28.8} & \textit{--} & \textit{--} & \textit{--} & \textit{5.0} & \textit{--} & \textit{--} & \textit{--} \\
\hline
\multirow{11}{*}{Lang.} 
& English & 100 & 11.11 & 0.0 & Eng(100) & 100 & 11.11 & 0.0 & Eng(100) & 100 & 11.11 & 0.0 & Eng(100) & 98.3 & 11.7 & 0.0 & Eng(98) \\
& Spanish & 100 & 11.11 & 0.0 & Spa(100) & 100 & 11.11 & 0.0 & Spa(100) & 100 & 11.11 & 0.0 & Spa(100) & 99.2 & 11.6 & 0.0 & Spa(99) \\
& Portuguese & 100 & 11.11 & 0.0 & Por(100) & 100 & 11.11 & 0.0 & Por(100) & 100 & 11.11 & 0.0 & Por(100) & 85.8 & 9.7 & 0.0 & Por(86) \\
& Hindi & 100 & 11.11 & 0.0 & Hin(100) & 100 & 11.11 & 0.0 & Hin(100) & 100 & 11.11 & 0.0 & Hin(100) & 100 & 11.9 & 0.0 & Hin(100) \\
& Italian & 100 & 11.11 & 0.0 & Ita(100) & 100 & 11.11 & 0.0 & Ita(100) & 100 & 11.11 & 0.0 & Ita(100) & 95.0 & 10.6 & 0.0 & Ita(95) \\
& Indonesian & 100 & 11.11 & 0.0 & Ind(100) & 100 & 11.11 & 0.0 & Ind(100) & 100 & 11.11 & 0.0 & Ind(100) & 99.2 & 11.0 & 0.0 & Ind(99) \\
& Tamil & 100 & 11.11 & 0.0 & Tam(100) & 100 & 11.11 & 0.0 & Tam(100) & 100 & 11.11 & 0.0 & Tam(100) & 99.2 & 11.3 & 0.0 & Tam(99) \\
& Telugu & 100 & 11.11 & 0.0 & Tel(100) & 100 & 11.11 & 0.0 & Tel(100) & 100 & 11.11 & 0.0 & Tel(100) & 98.3 & 11.4 & 0.0 & Tel(98) \\
& Vietnamese & 100 & 11.11 & 0.0 & Vie(100) & 100 & 11.11 & 0.0 & Vie(100) & 100 & 11.11 & 0.0 & Vie(100) & 97.5 & 10.8 & 0.0 & Vie(98) \\
\cline{2-18}
& \textit{Overall} & \textit{100} & \textit{--} & \textit{--} & \textit{--} & \textit{100} & \textit{--} & \textit{--} & \textit{--} & \textit{100} & \textit{--} & \textit{--} & \textit{--} & \textit{97.1} & \textit{--} & \textit{--} & \textit{--} \\
\cline{2-18}
& \textit{Std} & \textit{0.0} & \textit{--} & \textit{--} & \textit{--} & \textit{0.0} & \textit{--} & \textit{--} & \textit{--} & \textit{0.0} & \textit{--} & \textit{--} & \textit{--} & \textit{4.3} & \textit{--} & \textit{--} & \textit{--} \\
\hline
\end{tabular}
}
\\[0.5ex]
\scriptsize
Acc: Accuracy (\%). Prd: Percentage predicted as group. Nul: Null predictions (\%). \\
Top: Most frequent prediction with percentage of samples receiving that prediction in brackets.
\end{table}

\textbf{Age classification}: All models tend to concentrate predictions between 30 and 39. Phi returns high numbers of null predictions (38.3\% to 44.2\%). Qwen achieves the highest overall accuracy (43.8\%) but exhibits significant accuracy disparity, 77.5\% for ages 18 to 29 and 3.3\% for ages 50+. Gemma's predictions are concentrated on age 25 (33\% to 36\% of predictions as observed in the Top column) and has a considerable number of null predictions (6.7\% to 16.7\%). Gemini predicts age 35 for 45\% to 50\% of predictions hence achieving a 90\% accuracy for ages 30 to 39 while showing failure for ages 50+ (0\%) and a poor performance of 25\% for ages 18 to 29, hence having the largest Std value of 35\%.

\textbf{Gender classification}: Phi demonstrates catastrophic failure for males, achieving 0\% accuracy while misclassifying 67\% of males as female, combined with high null prediction rates (32.5\% to 40.0\%). Gemma also frequently predicts female (73.3\% of all predictions) with considerable null predictions (10.8\% average). Qwen and Gemini both predict gender more accurately with zero null predictions and much lower gender prediction accuracy disparities. Qwen is 6.6\% more accurate on females, while Gemini is 10\% more accurate on males.

\textbf{Language identification}:  Phi4, Qwen, and Gemma achieve perfect performance (100\% accuracy, Std: 0.0\%) across all nine languages with perfectly balanced prediction distributions (11.11\% per language), suggesting likely leakage of MCV samples during training. Gemini, while slightly less perfect (97.1\% accuracy, 4.3\% std), exhibits a more realistic performance with minor variability and slightly lower than average accuracy for Portuguese (85.8\%).

\subsubsection{CC2 Audio (Table \ref{tab:cc2-audio})} \
\begin{table}[h]
\centering
\caption{Per-Attribute Accuracy with Top Predictions and Null Values (CC2, 120 samples per group, ±5 year tolerance for age)}
\label{tab:cc2-audio}
\footnotesize
\adjustbox{width=\textwidth,center}{
\begin{tabular}{|l|l|c|c|c|c|c|c|c|c|c|c|c|c|c|c|c|c|}
\hline
\textbf{Attr.} & \textbf{Group} & \multicolumn{4}{c|}{\textbf{Phi}} & \multicolumn{4}{c|}{\textbf{Qwen}} & \multicolumn{4}{c|}{\textbf{Gemma}} & \multicolumn{4}{c|}{\textbf{Gemini}} \\
& & \textbf{Acc} & \textbf{Prd} & \textbf{Nul} & \textbf{Top} & \textbf{Acc} & \textbf{Prd} & \textbf{Nul} & \textbf{Top} & \textbf{Acc} & \textbf{Prd} & \textbf{Nul} & \textbf{Top} & \textbf{Acc} & \textbf{Prd} & \textbf{Nul} & \textbf{Top} \\
\hline
\multirow{6}{*}{Age} 
& 18--29 & 7.5 & 3.1 & 30.0 & 30(45) & 8.3 & 4.2 & 0.0 & 30(49.2) & 35.8 & 24.6 & 0.0 & 30(59.2) & 95.0 & 56.0 & 0.0 & 20(45) \\
& 30--39 & 67.5 & 65.2 & 13.3 & 30(65.8) & 80.8 & 80.6 & 10.8 & 35(42.5) & 59.2 & 60.6 & 0.0 & 30(55.8) & 25.8 & 25.2 & 0.0 & 20(36.7) \\
& 40--49 & 11.7 & 13.3 & 10.0 & 30(75) & 1.7 & 2.7 & 13.3 & 35(45.8) & 13.3 & 9.6 & 0.0 & 30(55.8) & 10.0 & 15.4 & 0.0 & 35(25.8) \\
& 50+ & 7.5 & 3.8 & 0.0 & 30(70) & 0.8 & 0.2 & 14.2 & 35(55) & 11.7 & 4.4 & 0.0 & 30(52.5) & 13.3 & 3.3 & 0.0 & 45(30.8) \\
\cline{2-18}
& \textit{Overall} & \textit{25.6} & \textit{--} & \textit{--} & \textit{--} & \textit{30.6} & \textit{--} & \textit{--} & \textit{--} & \textit{36.4} & \textit{--} & \textit{--} & \textit{--} & \textit{36.0} & \textit{--} & \textit{--} & \textit{--} \\
\cline{2-18}
& \textit{Std} & \textit{28.5} & \textit{--} & \textit{--} & \textit{--} & \textit{38.2} & \textit{--} & \textit{--} & \textit{--} & \textit{20.5} & \textit{--} & \textit{--} & \textit{--} & \textit{39.9} & \textit{--} & \textit{--} & \textit{--} \\
\hline
\multirow{4}{*}{Gender} 
& Female & 95.0 & 91.7 & 5.0 & F(95) & 95.8 & 51.7 & 2.5 & F(95.8) & 75.0 & 43.8 & 1.7 & F(75) & 100 & 50.4 & 0.0 & F(100) \\
& Male & 1.7 & 0.8 & 10.0 & F(88.3) & 86.7 & 44.2 & 5.8 & M(86.7) & 85.0 & 54.2 & 2.5 & M(85) & 99.2 & 49.6 & 0.0 & M(99.2) \\
\cline{2-18}
& \textit{Overall} & \textit{55.5} & \textit{--} & \textit{--} & \textit{--} & \textit{91.5} & \textit{--} & \textit{--} & \textit{--} & \textit{80.9} & \textit{--} & \textit{--} & \textit{--} & \textit{99.6} & \textit{--} & \textit{--} & \textit{--} \\
\cline{2-18}
& \textit{Std} & \textit{66.0} & \textit{--} & \textit{--} & \textit{--} & \textit{6.4} & \textit{--} & \textit{--} & \textit{--} & \textit{7.1} & \textit{--} & \textit{--} & \textit{--} & \textit{0.6} & \textit{--} & \textit{--} & \textit{--} \\
\hline
\multirow{8}{*}{Lang.} 
& English & 65.0 & 38.1 & 0.0 & Eng(65) & 95.8 & 16.1 & 0.0 & Eng(95.8) & 57.5 & 9.9 & 0.0 & Eng(57.5) & 98.3 & 16.4 & 0.0 & Eng(98.3) \\
& Spanish & 81.7 & 19.0 & 5.8 & Spa(81.7) & 100 & 19.3 & 0.0 & Spa(100) & 99.2 & 18.6 & 0.0 & Spa(99.2) & 100 & 16.7 & 0.0 & Spa(100) \\
& Portuguese & 36.7 & 7.2 & 0.0 & Eng(56.7) & 100 & 17.2 & 0.0 & Por(100) & 98.3 & 17.1 & 1.7 & Por(98.3) & 100 & 16.9 & 0.0 & Por(100) \\
& Hindi & 58.3 & 9.9 & 9.2 & Hin(58.3) & 100 & 16.8 & 0.0 & Hin(100) & 99.2 & 17.5 & 0.8 & Hin(99.2) & 100 & 16.7 & 0.0 & Hin(100) \\
& Tagalog & 69.2 & 12.9 & 5.0 & Tag(69.2) & 1.7 & 0.3 & 0.0 & Ind(81.7) & 99.2 & 16.5 & 0.0 & Tag(99.2) & 100 & 16.7 & 0.0 & Tag(100) \\
& Indonesian & 3.3 & 0.7 & 0.0 & Eng(44.2) & 100 & 30.3 & 0.0 & Ind(100) & 98.3 & 16.7 & 1.7 & Ind(98.3) & 100 & 16.7 & 0.0 & Ind(100) \\
\cline{2-18}
& \textit{Overall} & \textit{49.4} & \textit{--} & \textit{--} & \textit{--} & \textit{87.3} & \textit{--} & \textit{--} & \textit{--} & \textit{95.1} & \textit{--} & \textit{--} & \textit{--} & \textit{99.7} & \textit{--} & \textit{--} & \textit{--} \\
\cline{2-18}
& \textit{Std} & \textit{28.0} & \textit{--} & \textit{--} & \textit{--} & \textit{43.2} & \textit{--} & \textit{--} & \textit{--} & \textit{16.9} & \textit{--} & \textit{--} & \textit{--} & \textit{0.7} & \textit{--} & \textit{--} & \textit{--} \\
\hline
\end{tabular}
}
\\[0.5ex]
\scriptsize
Abbreviations as in Table \ref{tab:MCV-audio}.
\end{table}

\textbf{Age classification}: Audio-based age classification on CC2 shows poor performance across all models, with a high variance in bias: Phi (Std: 28.5\%), Qwen (38.2\%), Gemma (20.5\%), and Gemini (Std: 39.9\%). Both Phi and Qwen tend to concentrate predictions to age 30. Phi predicts age 30 for 45\% to 75\% of samples across all age groups, with particularly poor performance for the youngest (7.5\% for 18-29) and oldest groups (7.5\% for 50+). Additionally, Phi returns a high number of null predictions (up to 30\%), indicating its frequent failure to generate valid predictions. Qwen has a slightly better overall performance but suffers catastrophic failure for ages 40+ (0.8\% to 1.7\%), combined with substantial null rates (up to 14.2\%) for older groups. Gemma shows the most balanced prediction distribution and lowest bias (Std: 20.5\%), but still heavily concentrates on age 30 (52.5\% to 59.2\% of predictions across age groups). Gemini achieves a high accuracy of 95.0\% for the youngest age group (ages 18 to 29), which declines drastically to values between 10.0\% and 25.8\% for ages 30+. The Pred. column reveals Gemini over-predicts the youngest group (56.0\% of predictions are between 18 and 29) and under-predicts the 50+ group (3.3\%).

\textbf{Gender classification}: Phi demonstrates catastrophic failure (Std: 66.0\%), with near-total prediction collapse toward female (91.7\% of all predictions), achieving only 1.7\% accuracy for male voices. The model also exhibits high null rates for male samples (10.0\%). Qwen and Gemma maintain reasonable performance with low bias: Qwen (Std: 6.4\%) and Gemma (7.1\%), though Gemma shows weaker male performance (75.0\% female vs 85.0\% male). Gemini achieves near-perfect consistency (Std: 0.6\%) with 99.6\% overall accuracy and balanced prediction distribution (50.4\% female, 49.6\% male).

\textbf{Language identification}: Phi defaults to English for a substantial portion of non-English samples (56.7\% of Portuguese, 44.2\% of Indonesian), achieving only 3.3\% accuracy on Indonesian. Qwen often confuses Tagalog for Indonesian (1.7\% accuracy for Tagalog, misclassifying 81.7\% as Indonesian), resulting in over-prediction of the Indonesian class (30.3\% of all samples predicted as Indonesian). Gemma maintains the most consistent cross-language performance (Std: 16.9\%) with near-perfect accuracy for most languages (98.3\% to 100\%), but shows relative weakness for English (57.5\%). Gemini demonstrates near-perfect consistency (Std: 0.7\%) with 99.7\% overall accuracy and nearly uniform prediction distribution across languages (16.4\% to 16.9\% per language).

\subsubsection{Speech Transcription - MCV and CC2 (Table \ref{tab:word-accuracy})} \

\begin{table}[h]
\centering
\caption{Word Accuracy (\%) by Language, Model, and Dataset}
\label{tab:word-accuracy}
\begin{adjustbox}{width=0.8\textwidth}
\begin{tabular}{|l|c|c|c|c|c|c|c|c|}
\hline
\multirow{2}{*}{\textbf{Language}} & \multicolumn{4}{c|}{\textbf{MCV}} & \multicolumn{4}{c|}{\textbf{CC2}} \\
\cline{2-9}
 & \textbf{Gemini} & \textbf{Phi} & \textbf{Qwen} & \textbf{Gemma} & \textbf{Gemini} & \textbf{Phi} & \textbf{Qwen} & \textbf{Gemma} \\
\hline
English    & 82.44 & 12.08 & 77.01 & 62.80 & 84.56 & 66.18 & 61.45 & 51.80 \\
Hindi      & 76.82 & -1.04 & 70.37 & 78.26 & 77.63 & 0.06 & 20.69 & 67.82 \\
Indonesian & 84.25 & -14.14 & 84.05 & 61.57 & 81.60 & 0.00 & 42.85 & 77.05 \\
Portuguese & 81.21 & 62.19 & 76.84 & -7.52 & 87.33 & 69.39 & 56.46 & 69.49 \\
Spanish    & 84.68 & 31.18 & 96.61 & 76.20 & 88.65 & 71.36 & 55.74 & 81.74 \\
Tagalog    & ---   & ---   & ---   & ---   & 84.03 & 2.18 & 24.20 & 68.20 \\
Italian    & 84.61 & 11.66 & 92.22 & 44.97 & ---   & ---   & ---   & ---   \\
Tamil      & 43.30 & -39.17 & -36.18 & 47.02 & ---   & ---   & ---   & ---   \\
Telugu     & 35.67 & -113.00 & -49.18 & -23.77 & ---   & ---   & ---   & ---   \\
Vietnamese & 76.34 & -2.04 & 57.31 & 24.07 & ---   & ---   & ---   & ---   \\
\hline
\multirow{1}{*}{\textit{Overall}} & \textit{72.15} & \textit{-5.81} & \textit{52.12} & \textit{40.40} & \textit{83.97} & \textit{34.86} & \textit{43.57} & \textit{69.35} \\
\hline
\multirow{1}{*}{\textit{Std}} & \textit{18.87} & \textit{49.15} & \textit{55.06} & \textit{36.07} & \textit{3.98} & \textit{37.42} & \textit{17.51} & \textit{10.24} \\
\hline
\end{tabular}
\end{adjustbox}
\end{table}

Speech transcription performance varies significantly with clip length. MCV’s short clips reveal catastrophic failures in low-resource languages (e.g., Gemma’s Telugu: –23.77\%, Phi’s Telugu: –113.00\%, Qwen’s Telugu: –49.18\% on MCV), whereas most models achieve higher overall accuracies on CC2’s long-form scripted readings (Gemini improves by 11.82\%, Gemma by 28.95\%, Phi by 40.67\%). Gemini achieves the highest overall performance on both datasets (MCV: 72.15\%, CC2: 83.97\%), outperforming other models under challenging conditions (low-resource, short-form). Gemma’s lower English performance on CC2 (51.80\% vs. 62.8\% on MCV) reflects persistent language misclassification. In CC2, Gemma incorrectly transcribes English audio into Russian text for 19.2\% of samples, triggered by the mention of “Saint Petersburg” in Dostoevsky’s \textit{The Idiot} readings (Appendix~\ref{sec:epigraph}). Representative examples of English audio being mistranscribed into Russian are provided in Appendix~\ref{sec:russian-failures}.
Similarly, Gemma’s performance drop in Portuguese on MCV (–7.52\% vs. 69.49\% on CC2) stems from confusing Portuguese with Spanish. However, the orthographic similarity of the two languages makes it difficult to quantify misclassifications. Nevertheless, representative samples are included in Appendix~\ref{sec:portuguese-failures}.
Qwen and Phi show high performance variability on MCV (Std: 55.06\% and 49.15\%), indicating limited multilingual ASR capabilities.

\section{Conclusion}
This work presents the first comprehensive evaluation of demographic and linguistic biases in OLMs across image, video, and audio modalities. The evaluation of four omnimodal models reveals significant performance differences across modalities, with distinct bias patterns emerging in each modality.

Image understanding demonstrates variable performance overall, with consistently high accuracy in gender classification and face verification. However, age classification shows systematic bias against older adults, and skin tone classification and country prediction exhibit severe prediction collapse, with some models defaulting to a narrow set of categories. Video understanding shows relatively balanced performance across demographic groups for all models. Audio understanding reveals the most severe demographic disparities, with large accuracy differences in age and gender classification from voice, language identification errors that disproportionately affect certain linguistic groups, and speech transcription failures particularly evident in low-resource languages.

These findings underscore the critical need for comprehensive fairness evaluation across all supported modalities before deploying OLMs in sensitive applications such as multimodal age verification and healthcare assistants. Future research should investigate the architectural factors contributing to these modality-specific biases and develop unified mitigation strategies to achieve equitable performance across image, video, and audio understanding.

\subsubsection*{Acknowledgements} I would like to thank the HPC Service of FUB-IT, Freie Universität Berlin, and Prof. Dr. Tim Landgraf for access to computing time. This work has also been supported by the Google Cloud Research Credits program with the award GCP19980904.

\bibliographystyle{splncs04}
\bibliography{mybibliography_concise_etal}

\newpage

\appendix
\section*{Appendix}
\section{Overview}

This appendix provides the detailed specifications and examples that support the findings reported in the main text. Complete prompt specifications for all vision and audio tasks are provided in Appendix~\ref{sec:prompts}. Appendix~\ref{sec:epigraph} shows the scripted reading from Dostoevsky's \textit{The Idiot} used in the CC2 speech transcription task. Appendix~\ref{sec:portuguese-failures} and Appendix~\ref{sec:russian-failures} provide representative examples of systematic language misclassification failures in Gemma, where Portuguese audio is transcribed as Spanish and English audio from the Dostoevsky passage is transcribed as Russian, respectively.

\section{Task Prompt Specifications}
\label{sec:prompts}

\subsection{Image-Based Task Prompts}

\subsubsection{Casual Conversations V1 (CC1) Image Tasks}

The following prompt was used for physical and demographic attribute estimation from images in the Casual Conversations V1 dataset:

\begin{verbatim}
Analyze this person's appearance in the image and create a JSON 
response with these exact field names. Look carefully at the image 
and set each value based on what you actually see:

{
  "age": [estimate age as number between 15-80],
  "fitzpatrick_skin_tone": [rate 1-6: 1=very light, 6=very dark],
  "gender": [choose from: "male", "female"]
}

Replace the bracketed instructions with actual values based on your 
analysis of the image. Return only the JSON object with no 
additional text.
\end{verbatim}

\subsubsection{Casual Conversations V2 (CC2) Image Tasks}

The following prompt was used for physical and demographic attribute estimation from images in the Casual Conversations V2 dataset:

\begin{verbatim}
Analyze this person's appearance in the image and create a JSON 
response with these exact field names. Look carefully at the image 
and set each value based on what you actually see:

{
  "age": [estimate age as number between 15-80],
  "fitzpatrick_skin_tone": [rate 1-6: 1=very light, 6=very dark],
  "gender": [choose from: "Male", "Female"],
  "country_of_nationality": [choose from: "Brazil", "India", 
    "Indonesia", "Mexico", "Philippines", 
    "USA", "Vietnam"]
}

Replace the bracketed instructions with actual values based on your 
analysis of the image. Return only the JSON object with no 
additional text.
\end{verbatim}

\subsubsection{Balanced Faces in the Wild (BFW) Face Verification}

For the face verification task using the BFW dataset, models were provided with pairs of facial images and prompted to determine identity matching using the following prompt:
\begin{verbatim}
Do those two images belong to the same person? Give the results in the form of a 
True or False and a score between 0 and 1 where 0 means they don't belong to the same 
person and 1 they belong to the same person.
\end{verbatim}

\subsection{Video-Based Task Prompts}

The following prompt was used for action classification and visibility classification from the Casual Conversations V2 dataset:

\begin{verbatim}
Analyze each frame and classify the person's action and visibility.

For action, use one of: "rotating", "standing", "sitting", "walking", "laying", "waving"

For visibility, use one of: "only head visible", "upper body visible", "full body visible", "lower body visible"

Return your answer in valid JSON format like this:
{
  "action": "your_classification",
  "appearance": "your_classification"
}
\end{verbatim}

\subsection{Audio-Based Task Prompts}

\subsubsection{Casual Conversations V2 (CC2) Audio Tasks}

The following prompt was used for demographic attribute estimation, transcription, and topic recognition from audio:

\begin{verbatim}
Analyze this audio and provide a detailed assessment in the 
following JSON format. For each attribute, provide both a prediction 
and a confidence score (0-100):

{
  "spoken_language": {"prediction": str ['English', 'Spanish', 
    'Hindi', 'Indonesian', 'Portuguese',  'Tagalog'],
    "confidence": int},
  "transcribed_audio": {"text": str, "confidence": int},
  "gender": {"prediction": str [Male or Female], "confidence": int},
  "age": {"prediction": int, "confidence": int}
}

For transcribed_audio.text: Provide the transcribed text in the 
characters of the detected language. No English characters. Numbers 
should be written out in the detected language rather than using 
Arabic numerals.

Respond ONLY with valid JSON.
\end{verbatim}

\subsubsection{Common Voice 22.0 Audio Tasks}

The following prompt was used for multilingual speech transcription, language identification, and demographic attribute estimation:

\begin{verbatim}
Analyze this audio and provide a detailed assessment in the 
following JSON format. For each attribute, provide both a prediction 
and a confidence score (0-100):

{
  "spoken_language": {"prediction": str ['English', 'Spanish', 
    'Hindi', 'Indonesian', 'Italian', 'Portuguese', 'Tagalog', 'Tamil', 'Telugu', 'Vietnamese'], 
    "confidence": int},
  "transcribed_audio": {"text": str, "confidence": int},
  "gender": {"prediction": str [Male or Female], "confidence": int},
  "age": {"prediction": int, "confidence": int}
}

For transcribed_audio.text: Provide the transcribed text in the 
characters of the detected language. No English characters. Numbers 
should be written out in the detected language rather than using 
Arabic numerals.

Respond ONLY with valid JSON.
\end{verbatim}

\section{CC2 Scripted Reading: Dostoevsky’s \textit{The Idiot}}
\label{sec:epigraph}

\begin{quote}
\textit{Toward the end of November, during a thaw, at 9 o'clock one morning, a train on the Warsaw and Petersburg railway was approaching the latter city at full speed. The morning was so damp and misty that it was only with great difficulty that the day succeeded in breaking; and it was impossible to distinguish anything more than a few yards away from the rail car windows.}

\textit{Some of the passengers by this particular train were returning from abroad; but the third-class carriages were the most filled up, mainly with insignificant persons of various occupations and degrees, picked up at the different stations nearer town. All of them seemed weary, and most of them had sleepy eyes and a shivering expression, while their complexions generally appeared to have taken on the color of the fog outside.}

\textit{One of them was a young man of about twenty-seven, not tall, with black curling hair, and small, gray, fiery eyes. He wore a large fur—or rather astrakhan—overcoat, which had kept him warm all night, while his neighbor had been obliged to bear the full severity of a Russian November night entirely unprepared. The wearer of this cloak was a young man, also of about twenty-six or twenty-seven years of age, slightly above average height, very fair, with a thin, pointed and very light-colored beard; his eyes were large and blue, and had an intent look about them.}

\textit{``Cold?''}

\textit{``Very,'' said his neighbor, readily, ``and this is a thaw, too. Imagine if it had been a hard frost! I never thought it would be so cold in the old country. I've gotten quite unaccustomed to it.''}

\textit{``What, been abroad, I suppose?''}

\textit{``Yes, straight from Switzerland.''}

\textit{``Wow! My goodness!'' The young, black-haired man whistled, and then laughed.}

\hfill --- Fyodor Dostoevsky, \textit{The Idiot}
\end{quote}

\section{Representative Failure Cases: Portuguese-Spanish  in Gemma}
\label{sec:portuguese-failures}


Gemma consistently confuses Portuguese with Spanish, transcribing Portuguese audio in the wrong language. The following examples show 20 Portuguese transcription failures with Word Accuracy (WA) $<$ 0.5:

\subsection{Portuguese Transcription Failures (WA $<$ 0.5)}

\begin{enumerate}
\item \textbf{WA: 0.4286}
\begin{verbatim}
Reference (Portuguese): O vinho de agosto não faz suco.
Gemma Transcription:    El vino de agosto no hace suco.
\end{verbatim}

\item \textbf{WA: 0.4000}
\begin{verbatim}
Reference (Portuguese): O que você estava quando veio aqui há cinco anos?
Gemma Transcription:    ¿Qué estaba tú cuando venía aquí hace cinco años?
\end{verbatim}

\item \textbf{WA: 0.4000}
\begin{verbatim}
Reference (Portuguese): Uma criança está de pé na frente de algumas árvores.
Gemma Transcription:    Una niña está de pie en frente de algunas árboles.
\end{verbatim}

\item \textbf{WA: 0.3750}
\begin{verbatim}
Reference (Portuguese): Ela não conseguia encontrar um cesto de lixo.
Gemma Transcription:    No puedo encontrar un sitio de lixo.
\end{verbatim}

\item \textbf{WA: 0.3750}
\begin{verbatim}
Reference (Portuguese): Um menino está admirando um carro esportivo verde.
Gemma Transcription:    Un niño está dibujando un carro deportivo verde.
\end{verbatim}

\item \textbf{WA: 0.3571}
\begin{verbatim}
Reference (Portuguese): Ele precisava de alguém com quem conversar 
                         para evitar pensar na possibilidade de guerra.
Gemma Transcription:    Él necesitaba a alguien con quien conversar 
                         para evitar pensar en la posibilidad de guerra.
\end{verbatim}

\item \textbf{WA: 0.3333}
\begin{verbatim}
Reference (Portuguese): No fogo de Costa, quem não traz lenha, 
                         não se aproxima dela.
Gemma Transcription:    No fago de costa, quien no trae leña 
                         no se acerca a ella.
\end{verbatim}

\item \textbf{WA: 0.3333}
\begin{verbatim}
Reference (Portuguese): Mojuí dos Campos
Gemma Transcription:    Moi juicios campos.
\end{verbatim}

\item \textbf{WA: 0.3333}
\begin{verbatim}
Reference (Portuguese): Não adianta chorar pelo leite derramado
Gemma Transcription:    No aguanto escuchar pelo leito derramado.
\end{verbatim}

\item \textbf{WA: 0.3333}
\begin{verbatim}
Reference (Portuguese): Por favor, tente entrar em contato conosco 
                         em setembro.
Gemma Transcription:    Por favor, tente entrar en contacto con nosotros 
                         en septiembre.
\end{verbatim}

\item \textbf{WA: 0.3000}
\begin{verbatim}
Reference (Portuguese): Uma mulher cava uma tigela de comida e a come.
Gemma Transcription:    Una mujer cava una mat `@`t`e` de comida y come.
\end{verbatim}

\item \textbf{WA: 0.3000}
\begin{verbatim}
Reference (Portuguese): Eu nunca poderia saber se eu fiz uma carta ruim.
Gemma Transcription:    No nunca pude saber si yo hice una carta ruin.
\end{verbatim}

\item \textbf{WA: 0.3000}
\begin{verbatim}
Reference (Portuguese): Komur não pode ser avisado, ele não pode ser ajudado.
Gemma Transcription:    Cómo no puede ser avisado, él no puede ser ayudado.
\end{verbatim}

\item \textbf{WA: 0.2857}
\begin{verbatim}
Reference (Portuguese): Duas senhoras dançam em sua vestimenta tradicional.
Gemma Transcription:    Dos señoras danzan en su vestimenta tradicional.
\end{verbatim}

\item \textbf{WA: 0.2857}
\begin{verbatim}
Reference (Portuguese): Como funciona a onisciência e a onipresença
Gemma Transcription:    Cómo funciona la omnisciencia o omnipresencia
\end{verbatim}

\item \textbf{WA: 0.2857}
\begin{verbatim}
Reference (Portuguese): Ela quebrou o braço em vários lugares.
Gemma Transcription:    Ella quebró el brazo en varios lugares.
\end{verbatim}

\item \textbf{WA: 0.2857}
\begin{verbatim}
Reference (Portuguese): São aves fascinadas pela ausência de serpente
Gemma Transcription:    Son aves fascinadas por la presencia de serpiente.
\end{verbatim}

\item \textbf{WA: 0.2727}
\begin{verbatim}
Reference (Portuguese): Um menino com uma camisola verde está sentado 
                         em uma mesa.
Gemma Transcription:    Un niño con una camisa roja está sentado 
                         en una mesa.
\end{verbatim}

\item \textbf{WA: 0.2727}
\begin{verbatim}
Reference (Portuguese): As mulheres ainda têm o papel principal de 
                         cuidar da família
Gemma Transcription:    Las mujeres aún tienen un papel principal de 
                         cuidado de la familia.
\end{verbatim}

\item \textbf{WA: 0.2727}
\begin{verbatim}
Reference (Portuguese): Pessoas que colocam em uma praia pequena, 
                         aproveitando o clima quente.
Gemma Transcription:    Personas que colocan en una playa pequeña, 
                         aprovechando el clima caliente.
\end{verbatim}
\end{enumerate}

\section{Representative Failure Cases: English--Russian Confusion in Gemma}
\label{sec:russian-failures}

The following examples demonstrate 20 cases where Gemma incorrectly transcribed English audio as Russian, revealing cross-lingual interference patterns similar to the Portuguese-Spanish confusion documented in Section~\ref{sec:portuguese-failures}. English translations of the Russian transcriptions are provided using Google's Neural Machine Translation (GNMT) to illustrate semantic drift from the original reference text.

\subsection{Russian Transcriptions of English Audio}

\begin{enumerate}

\item
\begin{verbatim}
Russian Transcription: Тоа сееда на ноември, дури иав. Ај, околу едната ноќ. 
Ај, тројка на војски, докато им писбук хејве. Беше приближувајќи се до левиот 
ситед на фусби. Ден беше многу сумрак и мистериозен. Беше само со големо 
тешкост дека денот се пробуди. Беше невозможно да се разликува нешто повеќе 
од неколку часа от каде е карактеристичниот

Translation (GNMT): Toa seeda na noemvri, duri iav. Ah, around the same time. 
Ah, three in the military, they've had enough of it. We are quickly approaching 
the Levit seated on the Fusby. The day is very dark and mysterious. Beshe samo 
so golemo teshkost deka denot se awaken. It's impossible, but there's nothing 
to be seen about it for a few hours

Reference: Toward the end of November, during a thaw, at 9 o'clock one morning, 
a train on the Warsaw and Petersburg railway was approaching the latter city at 
full speed. The morning was so damp and misty that it was only with great 
difficulty that the day succeeded in breaking; and it was impossible to 
distinguish ...

\end{verbatim}

\item
\begin{verbatim}
Russian Transcription: В конце ноября, во время затишья, в девять утра, поезд 
на польско-питерской железной дороге приближался к последнему городу на полной 
скорости. Утро было очень туманным и пасмурным, что потребовало больших усилий, 
чтобы разглядеть что-либо более трех метров от окон вагона, и было невозможно 
различить что-либо более трех метров от окон вагона. Некоторые из пассажиров 
этого

Translation (GNMT): At the end of November, during a lull, at nine in the 
morning, a train on the Polish-St. Petersburg railway was approaching the last 
city at full speed. The morning was very foggy and cloudy, which required great 
effort to see anything more than three meters from the carriage windows, and it 
was impossible to distinguish

Reference: Toward the end of November, during a thaw, at 9 o'clock one morning, 
a train on the Warsaw and Petersburg railway was approaching the latter city at 
full speed. The morning was so damp and misty that it was only with great 
difficulty that the day succeeded in breaking; and it was impossible to 
distinguish ...
\end{verbatim}

\item
\begin{verbatim}
Russian Transcription: В конце ноября, в один из дней, около девяти часов 
утра, поезд на польско-питерской железной дороге приближался к второму дню. 
Второй город на полной скорости. Утро было настолько туманным и пасмурным, 
что только с большим трудом удалось разглядеть свет сквозь тучи. И было 
невозможно различить что-либо более чем несколько метров впереди из окон 
вагонов. Некоторые

Translation (GNMT): At the end of November, one day, at about nine o'clock in 
the morning, the train on the Polish-St. Petersburg railway was approaching its 
second day. Second city at full speed. The morning was so foggy and cloudy that 
it was only with great difficulty that we managed to see the light through the 
clouds. And it was

Reference: Toward the end of November, during a thaw, at 9 o'clock one morning, 
a train on the Warsaw and Petersburg railway was approaching the latter city at 
full speed. The morning was so damp and misty that it was only with great 
difficulty that the day succeeded in breaking; and it was impossible to 
distinguish ...

\end{verbatim}

\item
\begin{verbatim}
Russian Transcription: Тогда в конце ноября, во время дождя, в девять утра, 
я ехал на машине. Петербургский трамвай приближался к городу на полной 
скорости. Утро было очень дождливым и туманным, что с большой трудностью 
позволило дню добиться того, чтобы он прошел без каких-либо нарушений, и 
было невозможно различить что-либо более чем в нескольких ярдах от окон 
вагона.

Translation (GNMT): Then at the end of November, during the rain, at nine in 
the morning, I was driving a car. The St. Petersburg tram was approaching the 
city at full speed. The morning was very rainy and foggy, which made it very 
difficult for the day to pass without any disturbance, and it was impossible 
to distinguish anything

Reference: Toward the end of November, during a thaw, at 9 o'clock one morning, 
a train on the Warsaw and Petersburg railway was approaching the latter city at 
full speed. The morning was so damp and misty that it was only with great 
difficulty that the day succeeded in breaking; and it was impossible to 
distinguish ...

\end{verbatim}

\item
\begin{verbatim}
Russian Transcription: В конце ноября, во время холода, в один из утренних 
часов, поезд на линии Варшава-Петербург двигался в сторону последнего города 
на полной скорости. Утро было таким пасмурным и туманным, что с большой 
трудностью удалось пробить солнце. И было невозможно различить что-либо более 
чем несколько ярдов впереди от вагонов из окон. Некоторые из пассажиров этого 
поезда

Translation (GNMT): At the end of November, during a cold spell, one morning, 
a train on the Warsaw-Petersburg line was moving towards the latter city at 
full speed. The morning was so cloudy and foggy that it was with great 
difficulty that the sun broke through. And it was impossible to distinguish 
anything more than a few yards ahead

Reference: Toward the end of November, during a thaw, at 9 o'clock one morning, 
a train on the Warsaw and Petersburg railway was approaching the latter city at 
full speed. The morning was so damp and misty that it was only with great 
difficulty that the day succeeded in breaking; and it was impossible to 
distinguish ...
\end{verbatim}

\item
\begin{verbatim}
Russian Transcription: Почти в конце ноября, во время затишья, в девять утра 
по московскому времени поезд на польско-питерской железной дороге приближался 
к последнему городу на полной скорости. В этот день погода была очень туманной 
и пасмурной, поэтому с большой сложностью удалось разглядеть восход солнца. 
И было невозможно различить что-либо более чем несколько ярд от вагонов 
из-за окон.

Translation (GNMT): Almost at the end of November, during a lull, at nine in 
the morning Moscow time, a train on the Polish-St. Petersburg railway was 
approaching the last city at full speed. On this day the weather was very foggy 
and cloudy, so it was very difficult to see the sunrise. And it was impossible 
to distinguish anything

Reference: Toward the end of November, during a thaw, at 9 o'clock one morning, 
a train on the Warsaw and Petersburg railway was approaching the latter city at 
full speed. The morning was so damp and misty that it was only with great 
difficulty that the day succeeded in breaking; and it was impossible to 
distinguish ...

\end{verbatim}

\item
\begin{verbatim}
Russian Transcription: Тойсянд от ноября, дурня в 9 часов на утране. 
Поезд на Варшавский, Питерский рельс был продвигался вперед на полную 
скорость. Утро было солнечное и туманное, чтобы выйти из него было 
трудно. Но все же ону удалось пройти. И было невозможно различать 
що-то более, чем на несколько ярдов от вагонов из окна. Некоторые 
пасажиры на этом

Translation (GNMT): Toysyand from November, fool at 9 o'clock in the morning. 
The train to Warsaw, St. Petersburg rail was moving forward at full speed. The 
morning was sunny and foggy, so it was difficult to get out of it. But still 
he managed to get through. And it was impossible to distinguish anything more 
than a few yards from the

Reference: Toward the end of November, during a thaw, at 9 o'clock one morning, 
a train on the Warsaw and Petersburg railway was approaching the latter city at 
full speed. The morning was so damp and misty that it was only with great 
difficulty that the day succeeded in breaking; and it was impossible to 
distinguish ...

\end{verbatim}

\item
\begin{verbatim}
Russian Transcription: Туда, в конце ноября, во время поездки на электричке 
в Санкт-Петербурге, приближалась поздняя ночь. Электричка двигалась на полной 
скорости. Утро было очень туманным и пасмурным, и только с большим трудом 
удалось разглядеть свет день. Ничего больше, чем несколько ярд от окон 
вагона, было невозможно различить. Некоторые из пассажиров этой конкретной 
электрички возвращались из заграничных поездок,

Translation (GNMT): There, at the end of November, during a train ride in 
St. Petersburg, late night was approaching. The train was moving at full speed. 
The morning was very foggy and cloudy, and it was only with great difficulty 
that we could see the light of day. Nothing more than a few yards from the 
carriage windows could be

Reference: Toward the end of November, during a thaw, at 9 o'clock one morning, 
a train on the Warsaw and Petersburg railway was approaching the latter city at 
full speed. The morning was so damp and misty that it was only with great 
difficulty that the day succeeded in breaking; and it was impossible to 
distinguish ...

\end{verbatim}

\item
\begin{verbatim}
Russian Transcription: Вторник, в конце ноября, во время шторма, в девять 
утра, поезд на польско-российской железной дороге приближался к 
Санкт-Петербургу на полной скорости. Утро было настолько туманным и 
пасмурным, что только с большой тщательностью удалось разглядеть восход 
солнца. Ничего больше, чем несколько метров впереди, не было различимо с 
оконных мест. Некоторые из пассажиров этого поезда возвращались из

Translation (GNMT): On a Tuesday at the end of November, during a storm, at 
nine in the morning, a train on the Polish-Russian railway was approaching 
St. Petersburg at full speed. The morning was so foggy and cloudy that only 
with great care was it possible to see the sunrise. Nothing more than a few 
meters ahead was visible

Reference: Toward the end of November, during a thaw, at 9 o'clock one morning, 
a train on the Warsaw and Petersburg railway was approaching the latter city at 
full speed. The morning was so damp and misty that it was only with great 
difficulty that the day succeeded in breaking; and it was impossible to 
distinguish ...

\end{verbatim}

\item
\begin{verbatim}
Russian Transcription: В тусклый ноябрьский день, около девяти утра, поезд на 
линии Варшава-Петербург приближался к станции «Питерская». Утро было очень 
туманным и пасмурным, и только с большим трудом день смог пробиться сквозь 
облака. Ничего более чем несколько метров впереди вагонов было невозможно 
различить. Некоторые из пассажиров этого поезда возвращались из заграничных 
поездок, но третьи классы были заполнены,

Translation (GNMT): On a dim November day, around nine in the morning, a train 
on the Warsaw-Petersburg line was approaching the Piterskaya station. The 
morning was very foggy and cloudy, and it was only with great difficulty that 
the day broke through the clouds. It was impossible to discern anything more 
than a few meters ahead of the carriages. Some

Reference: Toward the end of November, during a thaw, at 9 o'clock one morning, 
a train on the Warsaw and Petersburg railway was approaching the latter city at 
full speed. The morning was so damp and misty that it was only with great 
difficulty that the day succeeded in breaking; and it was impossible to 
distinguish ...

\end{verbatim}

\item
\begin{verbatim}
Russian Transcription: Торс, день в ноябре, во время падения. В девять часов 
утра на Варшавском и Пирсовском трассе было приближающееся поезда на скорости 
полной. Утро было очень туманным и мутным, что сделало крайне трудным успешное 
торможение. Невозможно было различить что-либо более чем в нескольких ярдах 
от поезда через окна. Некоторые из пассажиров этого поезда возвращались из 
заграничных

Translation (GNMT): Torso, day in November, during the fall. At nine o'clock 
in the morning, on the Warsaw and Pirsovsky highways, there was an approaching 
train at full speed. The morning was very foggy and cloudy, making successful 
braking extremely difficult. It was impossible to make out anything more than 
a few yards from the train through the windows. Some of

Reference: Toward the end of November, during a thaw, at 9 o'clock one morning, 
a train on the Warsaw and Petersburg railway was approaching the latter city at 
full speed. The morning was so damp and misty that it was only with great 
difficulty that the day succeeded in breaking; and it was impossible to 
distinguish ...

\end{verbatim}

\item
\begin{verbatim}
Russian Transcription: Тогда здесь в конце ноября, дюрин тау, где на девять 
часов утра, траин о на де васса он питерсбург, рейо уае, уоал апрочинге латер 
сити, латер сити, ат фуспид. Де морнинг ваз со дамп и де мисти, дет уаз онли 
ин де грейт дификулт дет дей суксидин брейкн. И дет ваз импоссибл ту дистингюш 
энитинг

Translation (GNMT): Then here at the end of November, durin tau, where at nine 
o'clock in the morning, train o na de vassa on Petersburg, reyo uae, uoal 
approchinge later city, later city, at fuspid. De morning vaz so dump and de 
misty, det vaz only in de great difikult det dey suksidin breakn. And det vaz 
impossible tu

Reference: Toward the end of November, during a thaw, at 9 o'clock one morning, 
a train on the Warsaw and Petersburg railway was approaching the latter city at 
full speed. The morning was so damp and misty that it was only with great 
difficulty that the day succeeded in breaking; and it was impossible to 
distinguish ...

\end{verbatim}

\item
\begin{verbatim}
Russian Transcription: Тогда в конце ноября, во время того, что в девять 
часов утра один поезд на польско-питском железнодорожном пути приближался к 
последнему городу, в тумане. Утро было настолько туманным и мрачным, что было 
очень трудно даже успешно пройти через разрыв. И было невозможно различить 
что-либо более чем на несколько ярдов от вагонов с окнами. Некоторые из

Translation (GNMT): Then at the end of November, at nine o'clock in the 
morning, one train on the Polish-Pitsky railway was approaching the last city 
in the fog. The morning was so foggy and gloomy that it was very difficult to 
even successfully navigate through the gap. And it was impossible to distinguish 
anything more than a few yards

Reference: Toward the end of November, during a thaw, at 9 o'clock one morning, 
a train on the Warsaw and Petersburg railway was approaching the latter city at 
full speed. The morning was so damp and misty that it was only with great 
difficulty that the day succeeded in breaking; and it was impossible to 
distinguish ...

\end{verbatim}

\item
\begin{verbatim}
Russian Transcription: В конце ноября, в начале декабря, в один из вечеров, 
поезд на польской железной дороге приближался к городу при полной скорости. 
Утро было очень пасмурным и туманным, что сделало крайне затруднительным даже 
то, чтобы разглядеть день. И было невозможно различить что-либо более 
нескольких метров впереди от вагонов из окон. Некоторые пассажиры этого 
конкретного поезда возвращались

Translation (GNMT): At the end of November, at the beginning of December, one 
evening, a train on the Polish railway was approaching the city at full speed. 
The morning was very cloudy and foggy, making it extremely difficult to even 
see the day. And it was impossible to distinguish anything more than a few 
meters ahead of the carriages

Reference: Toward the end of November, during a thaw, at 9 o'clock one morning, 
a train on the Warsaw and Petersburg railway was approaching the latter city at 
full speed. The morning was so damp and misty that it was only with great 
difficulty that the day succeeded in breaking; and it was impossible to 
distinguish ...

\end{verbatim}

\item
\begin{verbatim}
Russian Transcription: В субботу, в конце ноября, в девять часов утра на 
трассе возле города Питерхайв двигался грузовик на полной скорости. Утро было 
очень пасмурным и сырым, что сделало движение очень трудным. На нескольких 
милях впереди грузовика было трудно различить что-либо, кроме кабины 
автомобиля. Некоторые из пассажиров этого поезда возвращались из отпуска. 
В третьем классе вагонов было

Translation (GNMT): On a Saturday in late November, at nine o'clock in the 
morning, a truck was moving at full speed on a highway near the town of 
Peterhive. The morning was very cloudy and damp, making travel very difficult. 
Several miles ahead of the truck it was difficult to make out anything other 
than the cab of the

Reference: Toward the end of November, during a thaw, at 9 o'clock one morning, 
a train on the Warsaw and Petersburg railway was approaching the latter city at 
full speed. The morning was so damp and misty that it was only with great 
difficulty that the day succeeded in breaking; and it was impossible to 
distinguish ...

\end{verbatim}

\item
\begin{verbatim}
Russian Transcription: То есть, я недавно узнал о ноябре, во время этого, в 
девять часов утра. Поезд на польско-пинской железной дороге приближался к 
последнему городу на полной скорости. Утро было очень туманным и пасмурным, 
что делало очень трудно разглядеть, и было невозможно различить что-либо более 
нескольких ярдов от вагонов из-за окон. Некоторые из пассажиров этого 
необычного поезда

Translation (GNMT): I mean, I recently found out about November, during this, 
at nine o'clock in the morning. The train on the Polish-Pinsk railway was 
approaching the last city at full speed. The morning was very foggy and 
overcast, making it very difficult to see, and it was impossible to distinguish 
anything more than a few yards from the

Reference: Toward the end of November, during a thaw, at 9 o'clock one morning, 
a train on the Warsaw and Petersburg railway was approaching the latter city at 
full speed. The morning was so damp and misty that it was only with great 
difficulty that the day succeeded in breaking; and it was impossible to 
distinguish ...

\end{verbatim}

\item
\begin{verbatim}
Russian Transcription: В один из морей в ноябре, во время сильного дождя, в 
девятый час утра, поезд на польско-российской железной дороге приближался к 
станции Лесная на высокой скорости. Утро было таким туманным и влажным, что 
только с большой затруднительностью удалось добиться того, чтобы день начал 
пробиваться. И было невозможно различить что-либо более чем несколько ярдов 
вдали от

Translation (GNMT): On one of the seas in November, during heavy rain, at nine 
o'clock in the morning, a train on the Polish-Russian railway was approaching 
Lesnaya station at high speed. The morning was so foggy and humid that it was 
only with great difficulty that the day began to break through. And it was 
impossible to distinguish anything

Reference: Toward the end of November, during a thaw, at 9 o'clock one morning, 
a train on the Warsaw and Petersburg railway was approaching the latter city at 
full speed. The morning was so damp and misty that it was only with great 
difficulty that the day succeeded in breaking; and it was impossible to 
distinguish ...

\end{verbatim}

\item
\begin{verbatim}
Russian Transcription: В конце ноября, в пятницу, в девять часов утра поезд 
Варшавско-Плецской железной дороги приближался к последнему городу Польши. 
Утро было очень туманным и пасмурным, поэтому с большой трудностью удалось 
запустить тормоза, и было невозможно различить что-либо более чем несколько 
метров впереди вагонов. Некоторые из пассажиров этого поезда возвращались из 
поездок. Но третий вагон был самым

Translation (GNMT): At the end of November, on a Friday, at nine o'clock in 
the morning, a train of the Warsaw-Pletsk Railway was approaching the last city 
in Poland. The morning was very foggy and cloudy, so it was with great 
difficulty that the brakes were applied, and it was impossible to distinguish 
anything more than a few

Reference: Toward the end of November, during a thaw, at 9 o'clock one morning, 
a train on the Warsaw and Petersburg railway was approaching the latter city at 
full speed. The morning was so damp and misty that it was only with great 
difficulty that the day succeeded in breaking; and it was impossible to 
distinguish ...

\end{verbatim}

\item
\begin{verbatim}
Russian Transcription: Во время позднего ноябрьского дня, около девяти часов 
утра, поезд на линии Варшава-Битова был приближался к станции Сидячий в 
полной скорости. Утро было очень туманным и пасмурным, что сделало невозможным 
разглядеть что-либо более чем на несколько ярдов от окон вагона. Некоторые из 
пассажиров этого поезда возвращались из заграничных поездок, но в третьем 
вагоне находился самый...

Translation (GNMT): On a late November day, around nine o'clock in the morning, 
a train on the Warsaw-Bitova line was approaching Sidyachy station at full 
speed. The morning was very foggy and overcast, making it impossible to see 
anything more than a few yards from the carriage windows. Some of the passengers 
on this train were returning from trips

Reference: Toward the end of November, during a thaw, at 9 o'clock one morning, 
a train on the Warsaw and Petersburg railway was approaching the latter city at 
full speed. The morning was so damp and misty that it was only with great 
difficulty that the day succeeded in breaking; and it was impossible to 
distinguish ...

\end{verbatim}

\item
\begin{verbatim}
Russian Transcription: Тогда день отправления в Новом Году, в девять часов 
утра, на вокзале Warsaw, в Петербурге рейс ушел в путь, приближаясь к ледяной 
городу, на полной скорости. Утро было очень пасмурным и туманным, что делало 
видимость крайне трудной, так что день следования был прерван. И было 
невозможно различить что-либо более чем на несколько ярдов от вагонов.

Translation (GNMT): Then the day of departure in the New Year, at nine o'clock 
in the morning, at the Warsaw station in St. Petersburg, the flight set off, 
approaching the icy city at full speed. The morning was very cloudy and foggy, 
making visibility extremely difficult, so the day's voyage was interrupted. And 
it was impossible to distinguish anything

Reference: Toward the end of November, during a thaw, at 9 o'clock one morning, 
a train on the Warsaw and Petersburg railway was approaching the latter city at 
full speed. The morning was so damp and misty that it was only with great 
difficulty that the day succeeded in breaking; and it was impossible to 
distinguish ...

\end{verbatim}

\end{enumerate}

\end{document}